%% file: main.tex
\def\year{2022}\relax
\documentclass[letterpaper]{article} 
\usepackage{aaai22}  
\usepackage{times}  
\usepackage{helvet}  
\usepackage{courier}  
\usepackage[hyphens]{url}  
\usepackage{graphicx} 
\usepackage{natbib}  
\usepackage{caption} 
\DeclareCaptionStyle{ruled}{labelfont=normalfont,labelsep=colon,strut=off} 
\usepackage{algorithm}
\usepackage{algorithmic}

\usepackage{newfloat}
\usepackage{listings}
\floatstyle{ruled}
\newfloat{listing}{tb}{lst}{}
\floatname{listing}{Listing}

\usepackage{subfigure}
\usepackage{amsmath}
\usepackage{amssymb}
\usepackage{multirow, arydshln, booktabs}
\usepackage[table,xcdraw]{xcolor}
\usepackage{pifont}

\newcommand{\cmark}{\ding{51}}
\newcommand{\xmark}{\ding{55}}

\usepackage{authblk}
\makeatletter
\renewcommand\AB@affilsepx{\quad\protect\Affilfont} 
\makeatother

\nocopyright
\title{Adaptive Logit Adjustment Loss for Long-Tailed Visual Recognition}
\author[1]{Yan Zhao}
\author[2]{Weicong Chen}
\author[3]{Xu Tan}
\author[2]{Kai Huang}
\author[1]{Jihong Zhu\thanks{Corresponding author}}
\affil[1]{Tsinghua University}
\affil[2]{ByteDance}
\affil[3]{Microsoft}

\affiliations{
    
}

\begin{document}

\maketitle

\input{sections/0_abstract.tex}
\input{sections/1_introduction.tex}

\input{sections/2_related_work.tex}
\input{sections/3_method.tex}
\input{sections/4_experiments.tex}
\input{sections/5_conclusion.tex}

\newpage
\bibliography{ref} 
\end{document}

%% file: sections/0_abstract.tex
\begin{abstract}
Data in the real world tends to exhibit a long-tailed label distribution, which poses great challenges for the training of neural networks in visual recognition.
Existing methods tackle this problem mainly from the perspective of data quantity, i.e., the number of samples in each class. To be specific, they pay more attention to tail classes, like applying larger adjustments to the logit. 
However, in the training process, the quantity and difficulty of data are two intertwined and equally crucial problems. 
For some tail classes, the features of their instances are distinct and discriminative, which can also bring satisfactory accuracy;
for some head classes, although with sufficient samples, the high semantic similarity with other classes and lack of discriminative features will bring bad accuracy.
Based on these observations, we propose Adaptive Logit Adjustment Loss (ALA Loss) to apply an adaptive adjusting term to the logit. The adaptive adjusting term is composed of two complementary factors: 1) \textit{quantity factor}, which pays more attention to tail classes, and 2) \textit{difficulty factor}, which adaptively pays more attention to hard instances in the training process. The difficulty factor can alleviate the over-optimization on \textit{tail yet easy} instances and under-optimization on \textit{head yet hard} instances.
The synergy of the two factors can not only advance the performance on tail classes even further, but also promote the accuracy on head classes. 
Unlike previous logit adjusting methods that only concerned about data quantity, ALA Loss tackles the long-tailed problem from a more comprehensive, fine-grained and adaptive perspective.
Extensive experimental results show that our method achieves the state-of-the-art performance on challenging recognition benchmarks, including ImageNet-LT, iNaturalist 2018, and Places-LT. 
\end{abstract}

%% file: sections/1_introduction.tex
\setcounter{secnumdepth}{2}\section{Introduction}

\begin{figure}[t]
\begin{center}
    \includegraphics[width=1\linewidth]{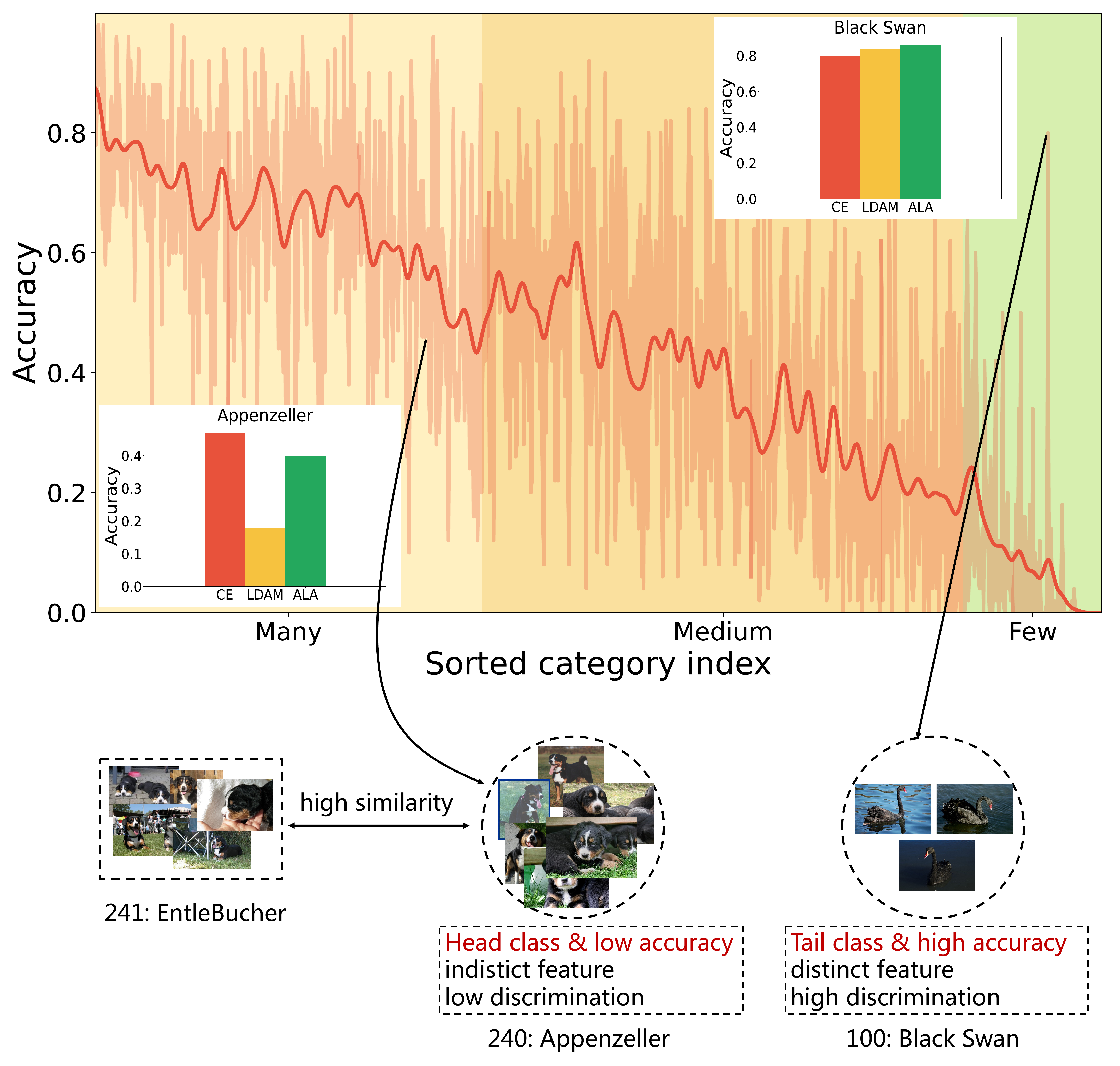}
\end{center}
\caption{Per-class accuracy of Cross Entropy (CE) method on ImageNet-LT dataset. The x-axis represents the class index sorted by the sample number. The y-axis shows the per-class accuracy. Best view in color and zoom in.}
\label{fig:intro_acc}
\vspace{-1.5em}
\end{figure}

With the development of deep learning, the computer vision community has witnessed the immense breakthrough of visual recognition on the classic benchmarks, such as ImageNet~\cite{russakovsky2015imagenet}, COCO~\cite{lin2014microsoft} and Places~\cite{zhou2017places}. 
In contrast to these artificially balanced datasets, real-world scenarios usually subject to a long-tailed label distribution. 
A few classes (head classes) contain most of the data, while most classes (tail classes) occupy relatively few samples~\cite{liu2019large,gupta2019lvis}. 
Unfortunately, confronted with such imbalanced distribution, the performance of these neural networks is found to degrade notably, especially on tail classes~\cite{cao2019learning,kang2019decoupling,liu2019large}.

Most existing long-tailed visual recognition methods address the problem by emphasizing the optimization on tail classes. These works can be roughly divided into three paradigms: re-sampling the training data~\cite{buda2018systematic, chawla2002smote, wallace2011class}, re-weighting the coefficients of loss formulations~\cite{menon2013statistical, cui2019class, ren2018learning} and adjusting the logit~\cite{cao2019learning, tan2020equalization, menon2020long}.
Data re-sampling increases the sampling rate for tail classes and decreases it for head classes.
Loss re-weighting guides the network to pay more attention to tail samples by up-weighting the tail classes and down-weighting the head classes.
Logit adjusting methods subtract a positive adjusting term from logit. This term is usually in reverse proportion to the frequency of each class, which encourages more optimization on tail classes.
They all tackle the long-tailed problem from the perspective of data quantity, sharing the same design philosophy: emphasizing more on tail classes, less on head classes.
However, according to our observations, \textit{data quantity is a necessary but insufficient condition.}

As shown in Figure~\ref{fig:intro_acc}, we plot the accuracy of each class on ImageNet-LT~\cite{liu2019large}, which is split by the number of training instances into few (1-20), medium (20-100) and many ($>$100) classes.
It is noticeable that although there is a certain correlation between accuracy and data quantity in general, it is not absolute from the perspective of each class.
For instance, class “Appenzeller” and “Black Swan” belong to the head and tail classes respectively. 
For “Black Swan”, despite comprising relatively few samples, it has high accuracy.
After searching all the bird group, we find that the characteristic of black swan is so distinct and discriminative, such as the black feather, the slender neck and the red beak, so that it can be easily distinguished.
However, for “Appenzeller”, even with sufficient samples, it still leads to a poor accuracy.
The indistinct property and high semantic similarity with other classes (such as “EntleBucher”) reduce its differentiability in the feature space, which greatly increases the risk of misclassification.
The above observations indicate that larger regularization is not needed for tail yet easy classes (like “Black Swan”), but is urgent for head yet hard classes (like “Appenzeller”).

To further shed light on the drawback of only focusing on data quantity, we also present specific comparison of accuracy between Cross Entropy (CE) and LDAM~\cite{cao2019learning}. LDAM is a prominent and effective method for long-tailed classification. It modifies the initial loss of CE by logit adjustment, but only from the perspective of data quantity.
As shown in the accuracy histogram of Figure~\ref{fig:intro_acc}, for the tail yet easy class “Black Swan” (in the upper right corner), the accuracy of CE is good enough (0.8). Although LDAM indeed has slight promotion (0.8 to 0.84), the gain is just marginal. In contrast, for the head yet hard class “Appenzeller” (in the lower left corner), CE achieves bad performance (0.47). Under this situation, LDAM deteriorates the performance severely (0.47 to 0.18).

Based on these observations, we propose a novel Adaptive Logit Adjustment Loss (ALA Loss), which encourages more regularization on not only tail classes (including all the instances in tail classes), but also hard instances (in both head and tail classes). 
The adjusting term of ALA Loss is composed of two complementary factors: 1) quantity factor, which pays more attention to tail classes; 2) difficulty factor, which adaptively regularizes more on hard instances in the training process by binding with the value of logit.
The synergy of the two factors can advance the performance on tail classes even further. More importantly, it mitigates the under-optimization on the hard samples of head classes, which promotes the accuracy on head classes at the same time. 
As we intended, compared to logit adjusting method LDAM, ALA Loss achieves better results on both the tail class “Black Swan” and the head class “Appenzeller” in Figure~\ref{fig:intro_acc}.

The contributions of our work can be summarized as follows: 
\begin{enumerate}
  \item We develop a novel Adaptive Logit Adjustment Loss (ALA Loss), which contains a quantity factor and a difficulty factor. ALA Loss works in a more comprehensive and fine-grained way. To be specific, previous methods only regularize more on tail classes. In comparison, ALA Loss takes both the quantity and difficulty of data into consideration, adjusting from the perspective of both class level and instance level. Supplemented with our difficulty factor, the over-optimization on tail yet easy and under-optimization on head yet hard instances can be efficaciously alleviated.
  \item We propose to adaptively apply regularization in the training process. Specifically, previous methods employ prior data quantity related to class frequency for logit adjustment. Our adjusting term takes a step further by adaptively choose which instances to regularize based on the value of the predicted logit. It can make the learning process more efficient, and boost the performance of both tail classes and hard samples.
  \item We conduct extensive and comprehensive experiments. ALA Loss shows consistent and significant improvements on three challenging large-scale long-tailed datasets, including ImageNet-LT, iNaturalist 2018 and Places-LT.
\end{enumerate}

%% file: sections/2_related_work.tex
\section{Related Works}

\subsection{Long-Tailed Classification}
Existing techniques for long-tailed classification mainly involves data re-sampling~\cite{chawla2002smote,han2005borderline,drumnond2003class}, loss modifying~\cite{cui2019class,khan2017cost,cao2019learning,ren2018learning,lin2017focal}, knowledge transferring~\cite{yin2019feature,liu2019large} and network structure designing~\cite{kang2019decoupling,zhou2020bbn,wang2020long}. 

As for data re-sampling, two common techniques are: over-sampling~\cite{chawla2002smote,han2005borderline} for tail classes and under-sampling~\cite{drumnond2003class} for head classes. As for loss modifying, it can be roughly classified into re-weighting based~\cite{cui2019class,ren2018learning,shu2019meta,lin2017focal} and logit adjusting based~\cite{cao2019learning,tan2020equalization,menon2020long} methods. Apart from the aforementioned strategies, knowledge transferring usually occurs from head to tail classes and the knowledge can be intra-class variance~\cite{yin2019feature} or semantic feature~\cite{liu2019large}. Recently, the methods of network structure designing also show promising success. \citet {kang2019decoupling} proposes a commonly used two-stage training strategy. \citet {xiang2020learning,zhou2020bbn,wang2020long} introduce multi-expert structure into long-tailed problem, sharing the same principle of divide-and-conquer. In this paper, we mainly focus on the simple but efficient logit adjusting losses, which can be easily integrated into other methods.

\subsection{Logit Adjustment}
Logit adjusting based loss is first proposed in face recognition~\cite{liu2016large,liu2017sphereface,wang2018additive,wang2018cosface,deng2019arcface}, which encourages larger inter-class margin and enforces extra intra-class compactness. LDAM~\cite{cao2019learning} introduces this idea to long-tailed recognition for the first time, and proposes a class-dependent adjusting term to enlarge margins for tail classes. Equalization Loss~\cite{tan2020equalization} applies logit adjustment to alleviate the overwhelmed discouraging gradients from head to tail classes in the field of object detection. Logit Adjust~\cite{menon2020long} analyzes from Fisher consistency and proposes a general form for logit adjustment. Among them, LDAM is a prominent and effective method for long-tailed classification, so we take it as the baseline of logit adjusting losses. 

%% file: sections/3_method.tex
\section{Method}\label{sec:3_method}
As shown in Figure~\ref{fig:intro_acc}, for long-tailed recognition, the quantity and difficulty of data are two intertwined and equally crucial part. Based on this observation, we propose Adaptive Logit Adjustment Loss (ALA Loss), which consists of two complementary factors, i.e., quantity factor and difficulty factor.

\subsection{Preliminary}\label{sec:method_preliminaries}
We first revisit the widely used softmax Cross-Entropy (CE) loss:
\begin{equation} \label{eq:celoss}
    \begin{aligned}
        \mathcal{L}_{CE}(y, f_{\theta}(x)) &= -\mathrm{log}(\sigma_{CE}(y, f_{\theta}(x))), \\
        \sigma_{CE}(y, f_{\theta}(x)) &= \frac{e^{f_{\theta}(x)[y]}}{\sum_{j=1}^C{e^{f_{\theta}(x)[j]}}},
    \end{aligned}
\end{equation}
where $x$ is the input instance, $y \in \{1, 2, \cdots, C\}$ is the corresponding target class. $C$ is the total number of classes.
$f_{\theta}(x)[j]$ is the predicted logit of the $j$-th class. $\sigma_{CE}(y, f_{\theta}(x))$ is the predicted probability of the classifier.

Next, we review logit adjusting losses in the field of long-tailed recognition. 
The common methods~\cite{cao2019learning, menon2020long, tan2020equalization} tackle the long-tailed classification by subtracting a positive adjusting term $\mathcal{A} \in \mathbb{R}^{C}$ from logit $f_{\theta}(x)$.
Therefore, the formulation of logit adjusting loss can be written as:
\begin{equation} \label{eq:laloss}
    \begin{aligned}
        \mathcal{L}_{LA}(y, f_{\theta}(x), \mathcal{A}) &= -\mathrm{log}(\sigma_{LA}(y, f_{\theta}(x), \mathcal{A})), \\
        \sigma_{LA}(y, f_{\theta}(x), \mathcal{A}) &= \frac{e^{f_{\theta}(x)[y]-\mathcal{A}[y]}}{\sum_{j=1}^{C}e^{f_{\theta}(x)[j] - \mathcal{A}[j]}},
    \end{aligned}
\end{equation}

Furthermore, the gradients of the loss $\mathcal{L}_{LA}$ on the logit $f_{\theta}(x)$ can be formulated as:
\begin{equation}
\label{eq:grad}
    \frac{\partial \mathcal{L}_{LA}}{\partial f_{\theta}(x)} = \left\{
    \begin{aligned}
    & \sigma_{LA}(y, f_{\theta}(x)[y], \mathcal{A}[y]) - 1, \,\, \text{for $j = y$}, \\
    & \sigma_{LA}(y, f_{\theta}(x)[j], \mathcal{A}[j]), \quad\quad\, \text{for $j \neq y$}, \\
    \end{aligned}
    \right.
\end{equation}

In previous works, $\mathcal{A}$ is only related to the data quantity. Specifically, it is a class-dependent term and negatively related to the number of samples in each class~\cite{cao2019learning,tan2020equalization,menon2020long}.
To further shed light on the effect of logit adjusting losses, we analyze it from the perspective of gradient. According to Equation~\eqref{eq:grad}, for the target class $y$, the gradient on logit is $\sigma_{LA}(y, f_{\theta}(x)[y], \mathcal{A}[y]) - 1$. 
For two samples with the same logit, denoting $f_{\theta}(x)[y_h]$ as the logit for the sample from head classes and $f_{\theta}(x)[y_t]$ for the one from tail classes, $f_{\theta}(x)[y_h] = f_{\theta}(x)[y_t]$. 
However, $\mathcal{A}[y_h] < \mathcal{A}[y_t]$, thus $\sigma_{LA}(y_h, f_{\theta}(x), \mathcal{A}) > \sigma_{LA}(y_t, f_{\theta}(x), \mathcal{A})$.  Because of $\sigma_{LA}(y, f_{\theta}(x)[y], \mathcal{A}) - 1 < 0$, thus $\left |  \frac{\partial \mathcal{L}_{LA}}{\partial f_{\theta}(x)} \right |_{y_h} < \left | \frac{\partial \mathcal{L}_{LA}}{\partial f_{\theta}(x)} \right |_{y_t}$. It means that, for two samples with the same logit, the one from tail classes will get a larger scale of gradient than that from head classes, making the model focus more on tail classes.

\subsection{ALA Loss}
Previous logit adjusting methods can effectively ameliorate the long-tailed situation from the perspective of data quantity. However, there are still some limitations. As shown in Figure~\ref{fig:intro_acc}, the over-optimization on tail yet easy and under-optimization on head yet hard samples are urgent issues to be settled. 
Therefore, we propose ALA Loss, whose form is the same as common logit adjusting losses shown in Equation~\eqref{eq:laloss}, but the adjusting term $\mathcal{A}$ is not only related to the data quantity but also correlated with the instance difficulty. ALA Loss designs $\mathcal{A}$ as the combination of a difficulty factor ($\mathcal{DF}$) and a quantity factor ($\mathcal{QF}$), as
\begin{equation}
    \label{eq:alaloss}
        \mathcal{A}^{ALA} = \mathcal{DF} \cdot \mathcal{QF},
\end{equation}
We will discuss the difficulty factor and quantity factor in detail, respectively.

\subsubsection{Difficulty Factor ($\mathcal{DF}$).}
$\mathcal{DF}$ is an instance-specific term, which aims to make the model pay more attention to hard instances. \textit{Since hard instances are those with worse predicted results, the design principle is that $\mathcal{DF}$ should be negatively related to the target prediction}.
Predicted logit and probability can both be utilized as the signal to measure the difficulty. We empirically find logit works better. The reason behind it is two folds: 1) Due to softmax, the predicted probability is sharper compared with the corresponding logit, which will lead to a over-large or over-small adjusting term. 2) The predicted logit have the same form with the original logit to be adjusted, which is more consistent and coherent.

However, as the value range of logit is unknown, it is hard to design the specific formulation of $\mathcal{DF}$. Therefore, we restrict $f_{\theta}(x)$ to $\left [ -1, 1 \right ]$ by weight normalization and feature normalization following LDAM. 
Specifically, we make the following transformations to $f_{\theta}(x_{i})$. $x_i$ is the $i$-th sample and it belongs to the $j$-th class:
\begin{equation}
\label{eq:trans}
    \begin{aligned}
    f_{\theta}(x_{i}) &:= W_j^Tx_i+b_j \\
        &:= \left \| W_j \right \| \left \| x_i \right \| \cos{\theta_{ij}} \quad \text{// by setting $b_j=0$}  \\
        &:= \left \| x_i \right \| \cos{\theta_{ij}} \quad\quad \text{// by weight normalization} \\
        &:= \cos{\theta_{ij}} \quad\quad\quad\quad \text{// by feature normalization}
    \end{aligned}
\end{equation}
Taking above transformations and design principle into consideration together, $\mathcal{DF}$ is designed negatively related to the value of $\cos{\theta_{ij}}$.
At the same time, the value range of $\mathcal{DF}$ is restricted to $\left [ 0, 1 \right ]$.
Then the formulation of $\mathcal{DF}$ is designed as:
\begin{equation}
\label{eq:df}
    \mathcal{DF} = \frac{1-\cos \theta_{iy}}{2},
\end{equation}
Note that we detach the gradient of $\cos \theta_{iy}$.

To further understand $\mathcal{DF}$, we give a more intuitive interpreting way.
$\theta_{ij}$ denotes the angle between $x_i$ and $W_j$. 
Since $W_j$ is generally considered as the target center of the $j$-th class~\cite{deng2019arcface}, $\theta_{iy}$ represents the angle between $x_i$ and its target class center, which is preferred to be as small as possible. 
For easy samples, $\theta_{iy}$ tends to be small, and $\cos \theta_{iy}$ tends to be large, which is vice versa for hard samples. Following Equation~\eqref{eq:df}, hard samples get more regularization than easy samples by our $\mathcal{DF}$, which brings improvements for discriminative learning.

\subsubsection{Quantity Factor ($\mathcal{QF}$).}
$\mathcal{DF}$ effectively makes the model focus more on hard instances, including those in tail classes. However, it is not enough. 
A further consideration is that instances in the same class jointly contribute to learn the representation and determine the classifier boundary. Hard samples in head classes can benefit from other samples of the same class due to the large number of samples, while hard samples in tail classes benefit less due to the lack of data samples in tail classes.

Therefore, we design a class-dependent term $\mathcal{QF}$ to better combine with $\mathcal{DF}$, making the network focus more on tail classes. \textit{The design principle is similar with previous logit adjusting losses: $\mathcal{QF}$ should be negatively related to the date quantity.} 
However, different from the common power function ($1/ x^n$) used in other methods~\cite{cao2019learning, menon2020long}, we empirically find log function ($1 / \log (x+1)$) is a better selection, which applies stronger regularization for tail classes.

Denoting $S = \left \{ S_1, S_2, ..., S_C \right \}$ as a set of sample number for each class, $\mathcal{QF}$ is formulated as following:
\begin{equation}
\label{eq:qf}
    \mathcal{QF} = \frac{1}{\log (\frac{S_j}{\min S} + 1)},
\end{equation}
where $S_j$ is normalized by the minimum number of samples $\min S$ following LDAM. Obviously, $\mathcal{QF}$ is class-dependent, whose value is larger for tail classes with smaller $S_j$. 

\begin{figure}[t]
    \centering
    \begin{center}
        \includegraphics[width=1\linewidth]{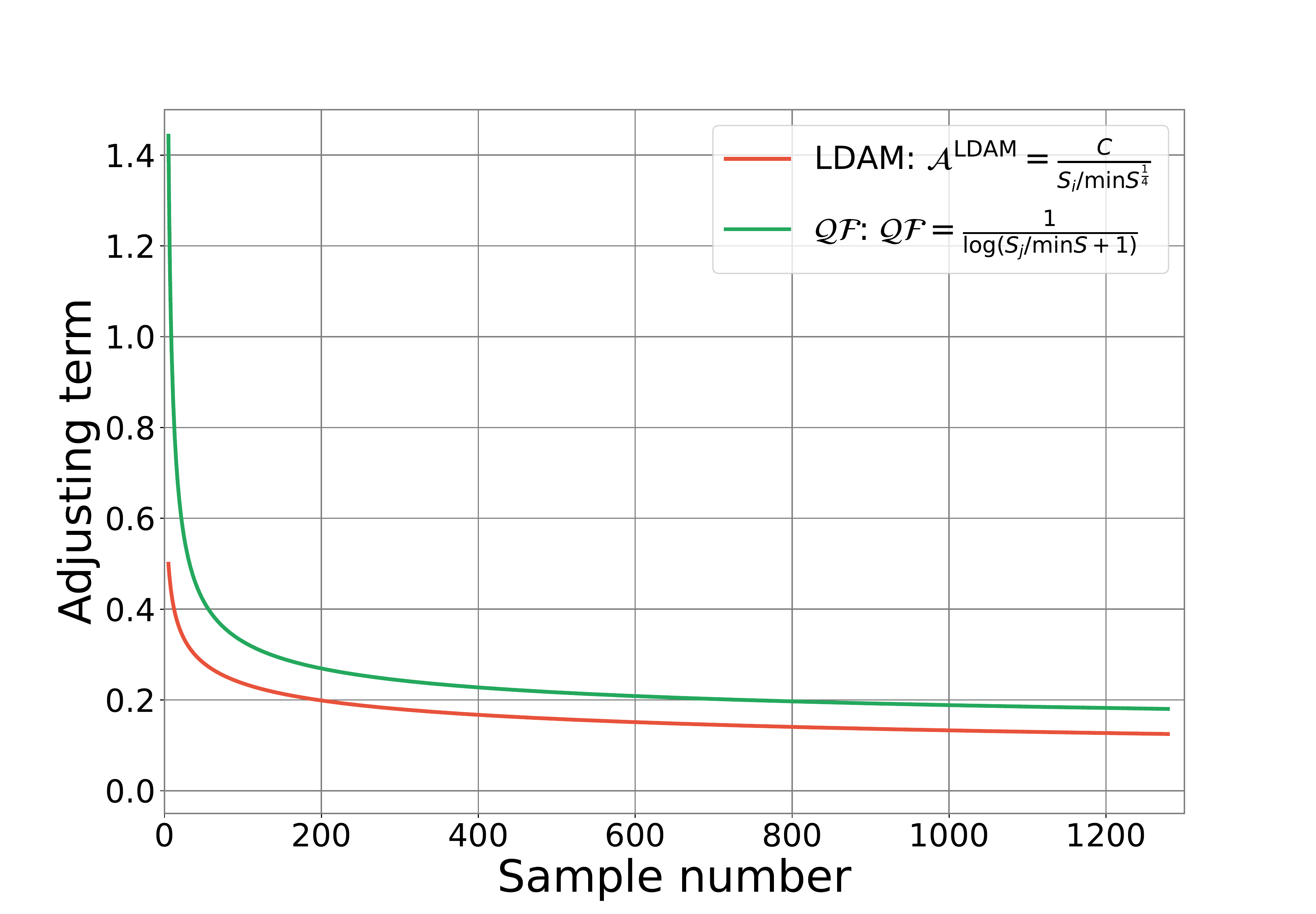}
    \end{center}
\caption{Comparison between LDAM and the $\mathcal{QF}$ term in ALA loss.}
\label{fig:method_qf}
\end{figure}

To further demonstrate the new function form of $\mathcal{QF}$ is essential, we intuitively show the comparison of our $\mathcal{QF}$ with LDAM in Figure~\ref{fig:method_qf}.
It can be clearly seen that $\mathcal{QF}$ assigns larger adjusting term to tail classes, which can boost the performance of them. The experimental results in Section~\ref{sec:ablation_quantitative} verify that $\mathcal{QF}$ is a more appropriate complementary term for $\mathcal{DF}$.

\subsubsection{Final Formation.}
Following LDAM~\cite{cao2019learning}, we only adjust the target logit, which means $\mathcal{A}[j] = 0$ when $j \neq y$, and we also re-scale the logit by the same constant $s$. Thus the final fomulation of our ALA Loss is
\begin{equation}
    \mathcal{L}_{ALA} = -\log{\frac{e^{s(f_{\theta}(x)[y] - \mathcal{A}^{ALA}[y])}}{e^{s(f_{\theta}(x)[y]  - \mathcal{A}^{ALA}[y])} + \sum_{j\neq y}^{C}e^{sf_{\theta}[j]}}}.
\label{eq:ala}
\end{equation}
Note that the design principle for $\mathcal{DF}$ is to be negatively related to the instance difficulty, and for $\mathcal{QF}$ is to be negatively related to the data quantity. Our formulations are designed following the motivation of being simple and practical. Other forms that conform to the principles are also suitable.

\subsection{Advantages over Previous Methods}
In summary, two appealing properties of ALA Loss make it stand out among previous logit adjusting losses.

1) \textit{ALA Loss is comprehensive and fine-grained.} 
The $\mathcal{DF}$ term in ALA Loss considers from the perspective of instance difficulty, which alleviates the over-optimization on tail yet easy and under-optimization on head yet hard instances.
For the $\mathcal{QF}$ term in ALA Loss, its design principle is the same as previous logit adjusting methods. However, we redesign the function to assign larger adjustments on tail classes, which makes it more suitable to integrate with $\mathcal{DF}$.

2) \textit{ALA Loss is adaptive.} 
Similar to previous methods~\cite{cao2019learning,menon2020long}, $\mathcal{QF}$ tackles the long-tailed recognition only considering the prior data quantity. In contrast, $\mathcal{DF}$ is related with the predicted logit, taking the dynamic status of training process into account. Consequently, our ALA Loss can adaptively focus on those poorly performing instances at present by giving them larger regularization, which is more rational and effective.

%% file: sections/4_experiments.tex
\section{Experiments}
\subsection{Dataset}

To evaluate the effectiveness and generality of our method, we conduct a series of experiments on three widely used large-scale long-tailed datasets: ImageNet-LT, iNaturalist 2018 and Places-LT.

\textbf{ImageNet-LT}. The ImageNet-LT~\cite{liu2019large} dataset is an artificially 
sampled subset of ImageNet-2012~\cite{deng2009imagenet}, with 115.8K images.
In this dataset, the overall number of classes is 1,000, while the maximum and minimum number of samples per class are 1,280 and 5, respectively.

\textbf{iNaturalist 2018}. The iNaturalist 2018~\cite{van2018inaturalist} dataset
is a real-world imbalanced dataset, with 437.5K images.
The overall number of classes is 8,142, with the maximum and minimum number of samples per class as 1,000 and 2, respectively.

\begin{table*}[ht]
\centering
\begin{tabular}{l|ccc|c|ccc|c}
\toprule
   \multirow{2}{*}{\textbf{Method}} &\multicolumn{4}{c|}{\textbf{Top-1 Accuracy @ R-50}} &\multicolumn{4}{c}{\textbf{Top-1 Accuracy @ X-50}} \\ \cmidrule{2-9} 
   &\textbf{Many} &\textbf{Medium} &\textbf{Few} &\textbf{All} &\textbf{Many} &\textbf{Medium} &\textbf{Few} &\textbf{All} \\ \midrule
   Cross Entropy $\dag$       &64.0 &33.8 &5.8  &41.6  &65.9 &37.5 &7.7 &44.4 \\
   Focal Loss $\ddag$         &-    &-    &-    &-     &64.3 &37.1 &8.2 &43.7  \\
   OLTR $\ddag$               &-    &-    &-    &-     &51.0 &40.8 &20.8 &41.9 \\
   Decouple-LWS $\dag$        &57.1 &45.2 &29.3 &47.7  &60.2 &47.2 &30.3 &49.9 \\
   DisAlign $\dag$            &59.9 &49.9 &31.8 &51.3  &61.5 &50.7 &33.1 &52.6 \\ \midrule

   Casual Norm $\ddag$        &-    &-    &-    &-     &62.7 &48.8 &31.6 &51.8 \\
   Balanced softmax $\ddag$   &-    &-    &-    &-     &62.2 &48.8 &29.8 &51.4 \\
   PC softmax $\ddag$         &-    &-    &-    &-     &60.4 &46.7 &23.8 &48.9 \\
   LADE $\ddag$               &-    &-    &-    &-     &62.3 &49.3 &31.2 &51.9 \\
   Logit Adjust (loss)        &- &- &-    &51.0 &-     &-    &-    &-    \\
   LDAM + DRW $*$             &61.8 &47.2 &31.4 &50.7  &62.9 &47.5 &31.9 &51.3 \\
   
   \cmidrule{1-1}\cmidrule{2-9}
   ALA Loss          &62.4 &49.1 &35.7 &\textbf{52.4}  &64.1 &49.9 &34.7 &\textbf{53.3}\\
\bottomrule
\end{tabular}
\caption{\textbf{Top-1 accuracy on the test set of ImageNet-LT equipped with ResNet50 (R-50) and ResNeXt50 (X-50).}
The superscript $\dag$ denotes that the results are from MisAlign~\cite{zhang2021distribution}, $\ddag$ are from LADE~\cite{hong2021disentangling} and $*$ means our reproduced results. The lower part from Causal Norm to LDAM + DRW are the logit adjusting methods.
}
\label{tab:exp_imagenet}
\end{table*}

\begin{table}[ht]
\centering
\begin{tabular}{l|ccc|c}
\toprule
   \textbf{Method} &\textbf{Many} &\textbf{Medium} &\textbf{Few} &\textbf{All} \\ \cmidrule{1-5}
   Cross Entropy $\dag$     & 72.2 & 63.0   & 57.2 & 61.7 \\
   CB-Focal $\ddag$         & -    & -      & -    & 61.1 \\
   Decouple-LWS $\dag$      & 65.0 & 66.3 & 65.5 & 65.9   \\ 
   BBN $\dag$               & 49.4 & 70.8 & 65.3 & 66.3   \\ 
   DisAlign $\dag$          & -    & -    & -    & 67.8   \\ \midrule
   
   Casual Norm $\ddag$      & -    & -    & -    & 63.9   \\
   Balanced Softmax $\ddag$ & -    & -    & -    & 69.8   \\
   PC Softmax $\ddag$       & -    & -    & -    & 69.3   \\
   LADE $\ddag$             & -    & -    & -    & 70.0   \\ 
   LDAM + DRW $\ddag$       & -    & -    & -    & 68.0 \\ \midrule
   ALA Loss                 & 71.3 & 70.8 & 70.4 & \textbf{70.7} \\ 
\bottomrule
\end{tabular}
\caption{\textbf{Top-1 accuracy on the validation set of iNaturalist 2018 with ResNet-50.} $\dag$ indicates that the results are from MisAlign. $\ddag$ indicates that the results are from LADE.}
\label{tab:exp_iNat18}
\vspace{-0.5em}
\end{table}

\begin{table}[ht]
\centering
\begin{tabular}{l|ccc|c}
\toprule
   \textbf{Method} &\textbf{Many} &\textbf{Medium} &\textbf{Few} &\textbf{All} \\ \cmidrule{1-5}
   Cross Entropy $\dag$     & 45.7 & 27.3   & 8.2 & 30.2 \\
   Focal Loss $\dag$        & 41.1 & 34.8 & 22.4 & 34.6 \\
   Range Loss $\dag$        & 41.1 & 35.4 & 23.2 & 35.1 \\
   OLTR $\dag$              & 44.7 & 37.0 & 25.3 & 35.9 \\   
   Feature Aug $\dag$       & 42.8 & 37.5 & 22.7 & 36.4 \\
   Decouple-LWS $\dag$      & 40.6 & 39.1 & 28.6 & 37.6  \\ 
   DisAlign $\dag$          & 40.4 & 42.4 & 30.1 & 39.3 \\ \midrule
   Causal Norm $\ddag$      & 23.8 & 35.8 & 40.4 & 32.4 \\
   Balanced Softmax $\ddag$ & 42.0 & 39.3 & 30.5 & 38.6 \\
   PC Softmax $\ddag$       & 43.0 & 39.1 & 29.6 & 38.7 \\
   LADE $\ddag$             & 42.8 & 39.0 & 31.2 & 38.8 \\ \midrule
   ALA Loss                 & 43.9 & 40.1 & 32.9 & \textbf{40.1} \\ 
\bottomrule
\end{tabular}
\caption{\textbf{Top-1 accuracy on the test set of Places-LT with ResNet-152.} $\dag$ indicates that the results are from DisAlign. $\ddag$ indicates that the results are from LADE.}
\label{tab:exp_places}
\vspace{-1em}
\end{table}

\begin{table*}[ht]
   \centering
   \begin{tabular}{c|c|c|c|ccc|c}
   \toprule
   &\textbf{LDAM} &\textbf{$\mathcal{DF}$} &\textbf{$\mathcal{QF}$} &\textbf{Many} &\textbf{Medium} &\textbf{Few} &\textbf{All} \\ \midrule
   Cross Entropy (CE) &\xmark &\xmark &\xmark  & 65.9 & 37.5 & 7.7  & 44.4 \\ \midrule
   LDAM &\cmark &\xmark &\xmark  & 62.9 & 47.5 & 31.9 & 51.3 \\
   $\mathcal{DF}$ &\xmark &\cmark &\xmark  & 64.7 & 47.3 & 28.6 & 51.5 \\
   $\mathcal{QF}$ &\xmark &\xmark &\cmark  & 61.4 & 47.7 & 33.0 & 51.0 \\ \midrule
   $\mathcal{DF} \cdot \mathcal{A}^{LDAM}$ &\cmark &\cmark &\xmark  & 63.5 & 48.4 & 31.9 & 52.0 \\ 
   $\mathcal{DF} \cdot \mathcal{QF}$ (ALA Loss) &\xmark &\cmark &\cmark  & 64.1 & 49.9 & 34.7 & \textbf{53.3} \\
   \bottomrule
   \end{tabular}
\caption{\textbf{Ablation studies for ALA Loss.} 
Results on the test set of ImageNet-LT with ResNeXt-50. The first line is the results of Cross Entropy (CE); the last line is our ALA Loss.
\textbf{$\mathcal{DF}$} denotes the difficulty factor in Equation~\eqref{eq:df}.
\textbf{$\mathcal{QF}$} denotes the quantity factor in Equation~\eqref{eq:qf}.}
\label{tab:exp_ablation}
\end{table*}

\textbf{Places-LT}. The Places-LT~\cite{liu2019large} dataset is a long-tailed subset of the dataset Places~\cite{zhou2017places}, with 62.5K images. It consists of 365 categories, the samples of each class ranging from 5 to 4,980.

\subsection{Experimental Setting}
\textbf{Implementation Details.}
For ImageNet-LT, ResNet-50 and ResNeXt-50 ($32\textrm{x}4$d)~\cite{he2016deep} are adopted as backbones. And we mainly use ResNeXt-50 for ablation studies.
The batch size is set as 256 with an initial learning rate of 0.1 and a weight decay of $0.0005$.
For iNaturalist 2018, ResNet-50 is used as the backbone.
And the batch size is set as 512 with an initial learning rate of 0.2 and a weight decay of $0.0002$.
For Places-LT, we utilize ResNet-152 as the backbone and pretrain it on the full ImageNet dataset following~\cite{hong2021disentangling} for fair comparison.

All networks are trained on 2 Tesla V100 GPUs, 90 epochs for ImageNet-LT and iNaturalist 2018, while 30 epochs for Places-LT.
The scale factor $s$ in Equation~\eqref{eq:ala} is set to 30 by default.
And we use the same training strategies as LDAM~\cite{cao2019learning} .

\textbf{Evaluation Protocol.}
All networks are trained on the long-tailed training datasets, and then evaluated on the corresponding balanced validation or test datasets. 
Top-1 accuracy is used as the evaluation metric, in the form of percentages. 
In order to better analyze the performance on classes of different data frequency, we report the accuracy on four class subsets according to the number of training instances in each class: \textit{Many-shot} ($>$100), \textit{Medium-shot} (20-100), \textit{Few-shot} (1-20) and \textit{All} as in~\cite{liu2019large}. 

\subsection{Main Results}
In this section, we present the performance of our method and compare with previous state-of-the-art works.
The results on ImageNet-LT, iNaturalist 2018 and Places-LT are shown in Table~\ref{tab:exp_imagenet}, Table~\ref{tab:exp_iNat18} and Table~\ref{tab:exp_places} respectively.

\subsubsection{Experimental Results on ImageNet-LT.}
As shown in Table~\ref{tab:exp_imagenet}, we have the following observations: 
1) ALA Loss achieves superior performance than existing methods on both networks. Comparing with the state-of-the-art method DisAlign, ALA Loss gets 1.1\% accuracy gain on ResNet50 and 0.7\% gain on ResNeXt50. 
2) Comparing with other logit adjusting losses: Logit Adjust~\cite{menon2020long} and LDAM, ALA Loss gets better results on all the three subsets, showing the full range of advantages of our method. 
3) Unlike other methods that improve tail classes at the sacrifice of head classes, our method not only achieves better results on tail classes, but also improves significantly on head classes, and achieves comparable results to Cross Entropy (CE), owing to the proposed Difficulty Factor.

\subsubsection{Experimental Results on iNaturalist 2018.}
As shown in Table~\ref{tab:exp_iNat18}, ALA Loss achieves better results than other methods on this real-world long-tailed dataset, showing the effectiveness of our method. Remarkably, ALA Loss obtains nearly equal result on all three subsets, which is a rather ideal result for long-tailed visual recognition~\cite{wang2020long}. Most of the existing methods improve the results of tail classes at the expense of head classes, while our method takes both into consideration.

\subsubsection{Experimental Results on Places-LT.} 
As shown in Table~\ref{tab:exp_places}, ALA Loss again outperforms other methods, especially on many- and few-shot subsets. Compared with other logit adjusting methods: from Casual Norm to LADE, ALA Loss achieves promising results, with 0.9\% gain on the many- and 0.8\% gain on the medium-shot. For the few-shot, ALA Loss achieves the best result other than Casual Norm. Casual Norm applies the post-hoc logit adjustment in the phase of inference. It can boost the accuracy of tail, however at the expense of head classes. Quantity and Difficulty Factor both contribute to the improvements of the few-shot subset, while the results of many-shot subset are mainly attributed to the Difficulty Factor.

\subsection{Ablation Study}

\begin{figure*}[t]
   \centering
   \subfigure[Many-shot]{
      \begin{minipage}[b]{0.3\textwidth}
         \includegraphics[width=1\textwidth]{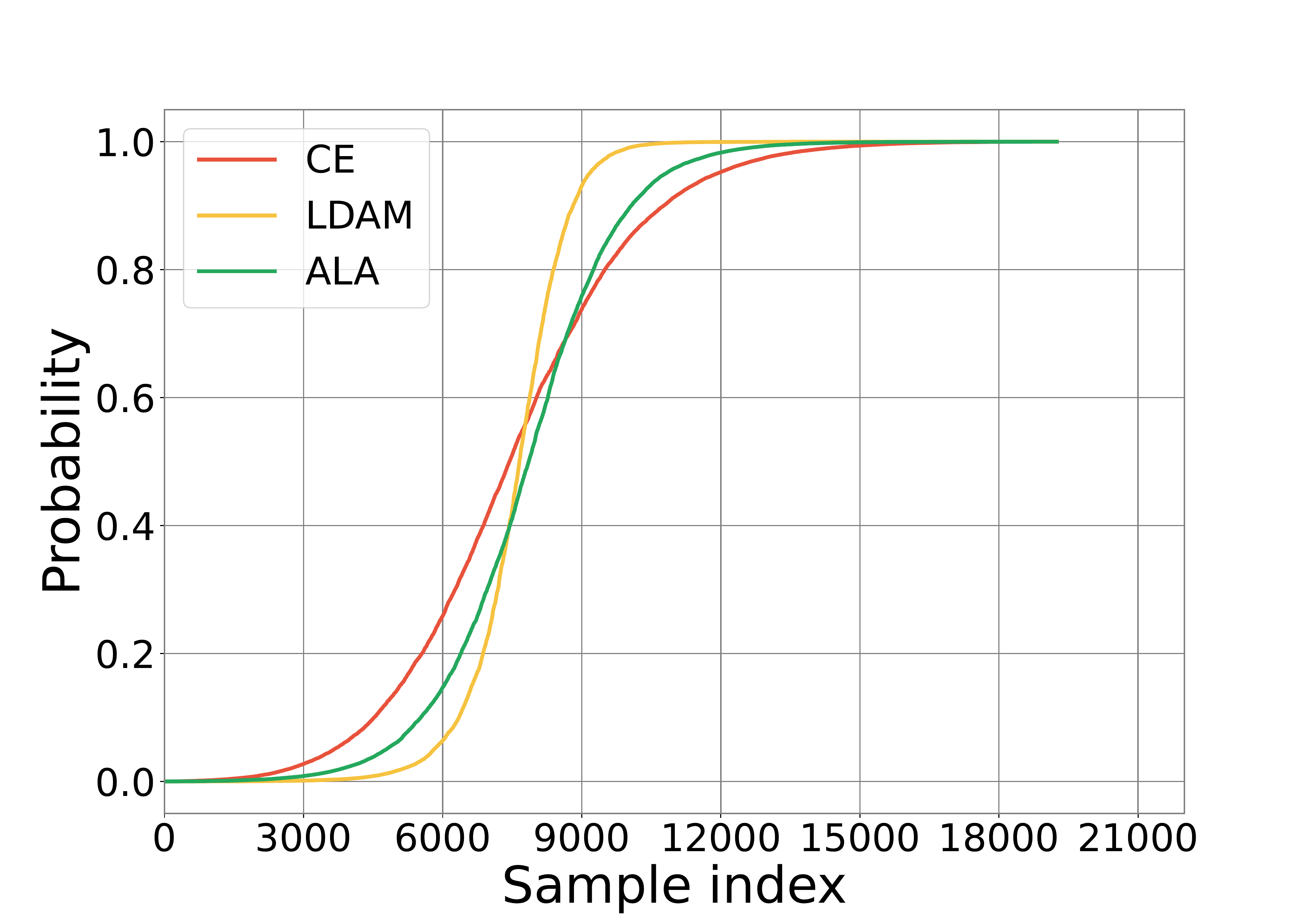}
      \end{minipage}
      \label{fig:exp_prob_many}
   }
   \subfigure[Medium-shot]{
      \begin{minipage}[b]{0.3\textwidth}
         \includegraphics[width=1\textwidth]{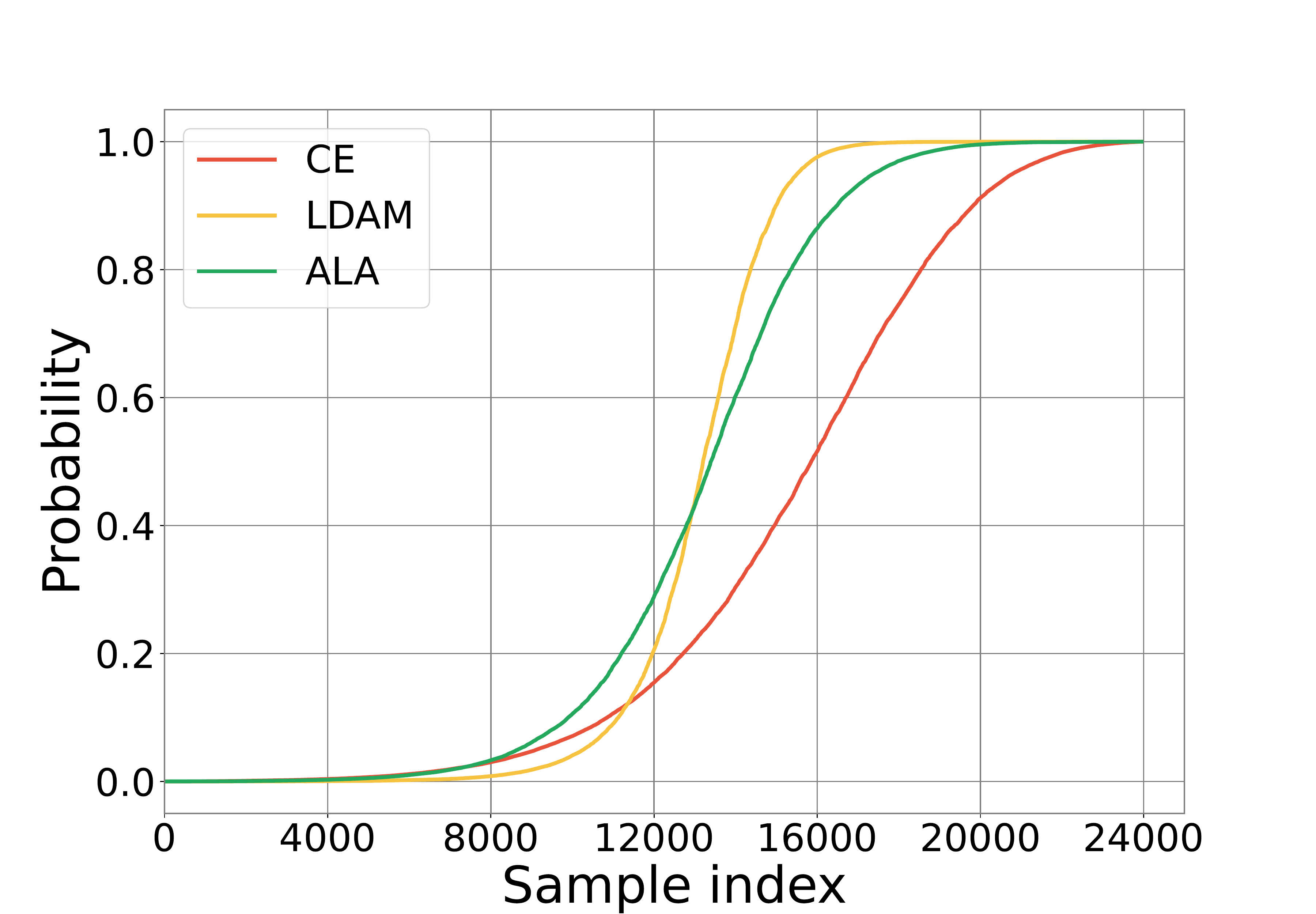}
      \end{minipage}
      \label{fig:exp_prob_med}
   }
   \subfigure[Few-shot]{
      \begin{minipage}[b]{0.3\textwidth}
         \includegraphics[width=1\textwidth]{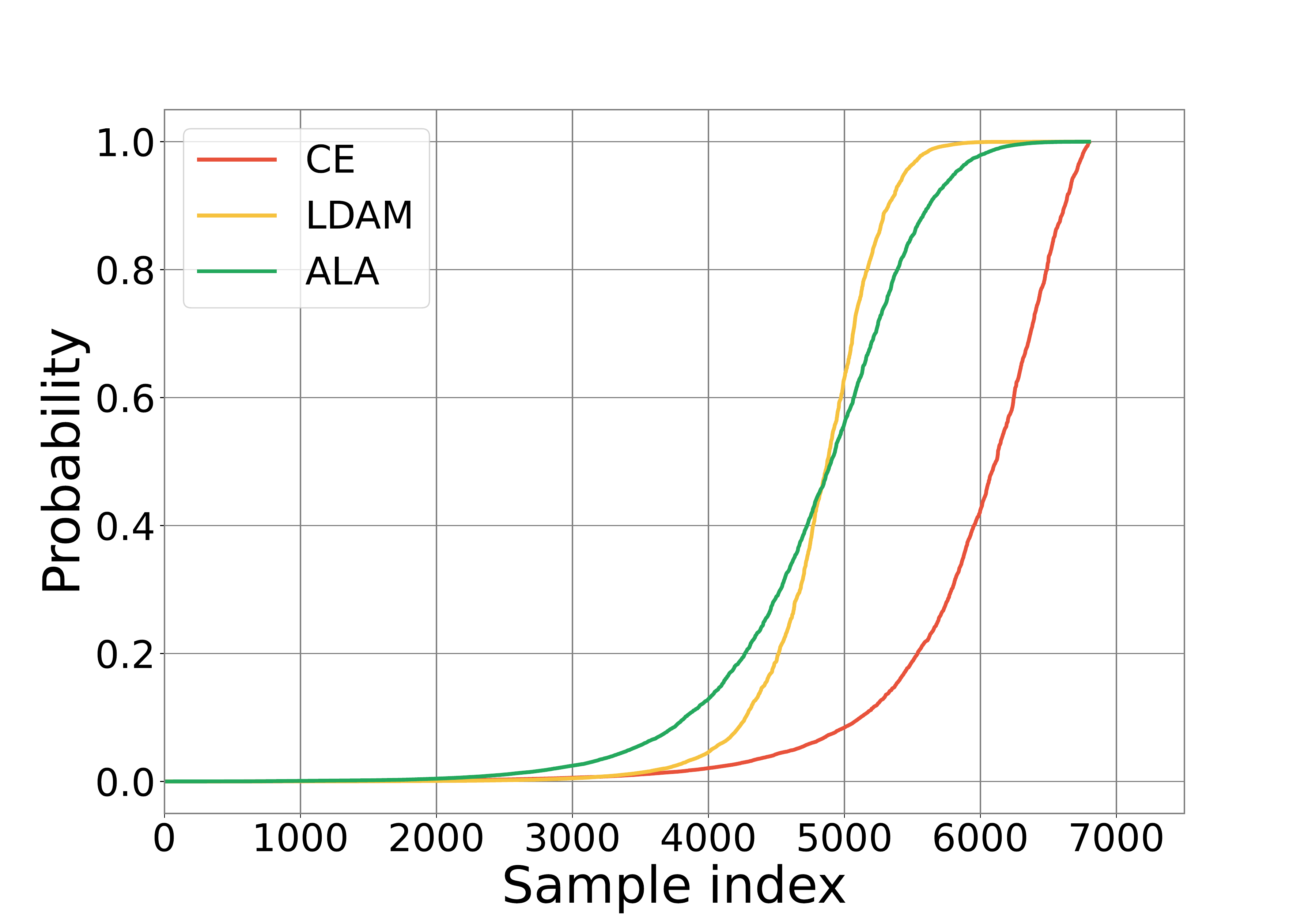}
      \end{minipage}
      \label{fig:exp_prob_few}
   }
   \caption{The predicted probability distributions for each class subset among CE, LDAM and ALA Loss. 
   The x-axis represents the sample index, which is sorted by the predicted probability. 
   There are more hard samples when the curve goes to the right.}
   \label{fig:exp_prob}
   \vspace{-1em}
\end{figure*}

\begin{figure*}[t!]
   \centering
   \subfigure[$\mathcal{DF}$]{
      \begin{minipage}[b]{0.3\textwidth}
         \includegraphics[width=1\textwidth]{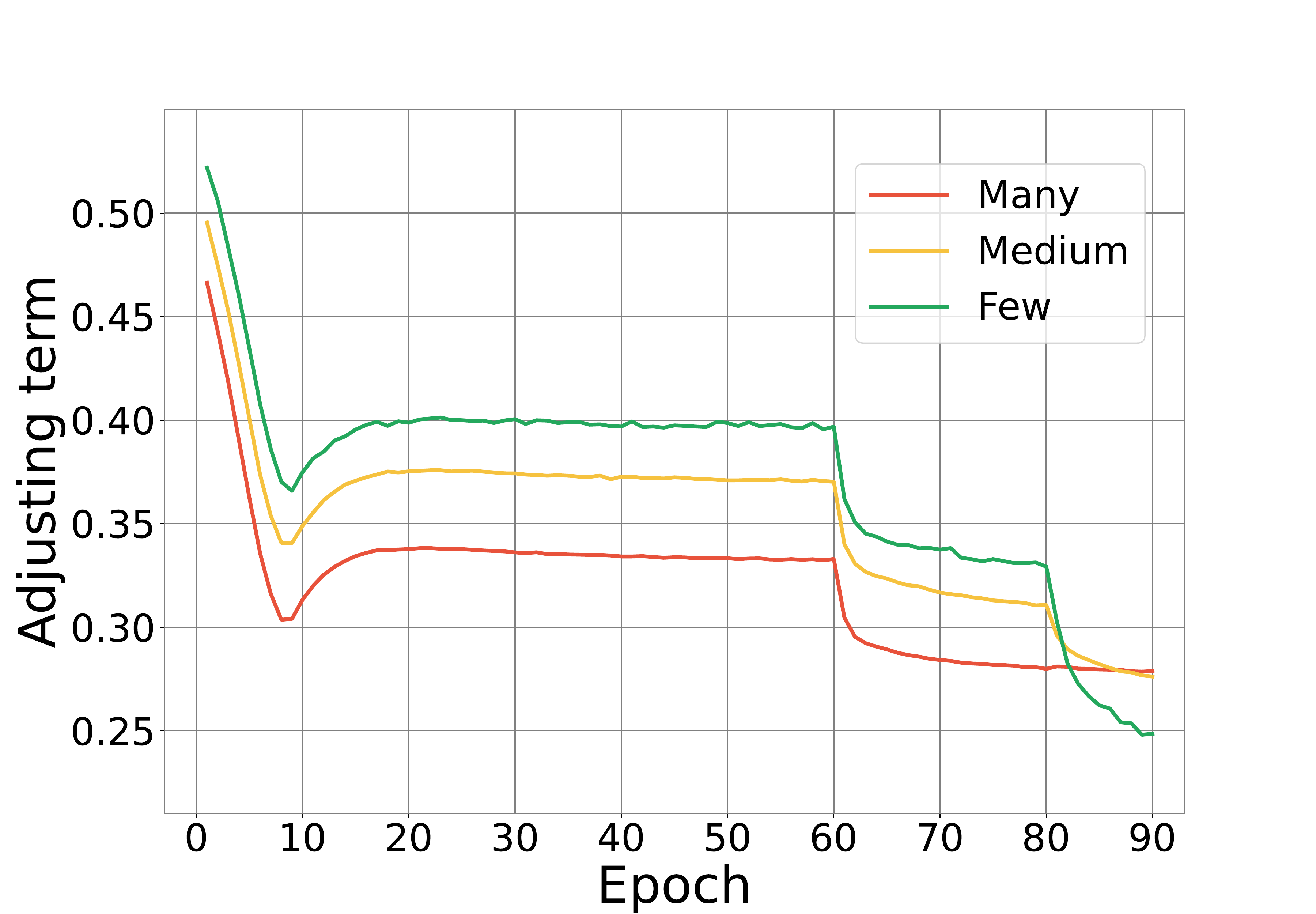}
      \end{minipage}
      \label{fig:exp_adjust_df}
   }
   \subfigure[$\mathcal{QF}$]{
      \begin{minipage}[b]{0.3\textwidth}
         \includegraphics[width=1\textwidth]{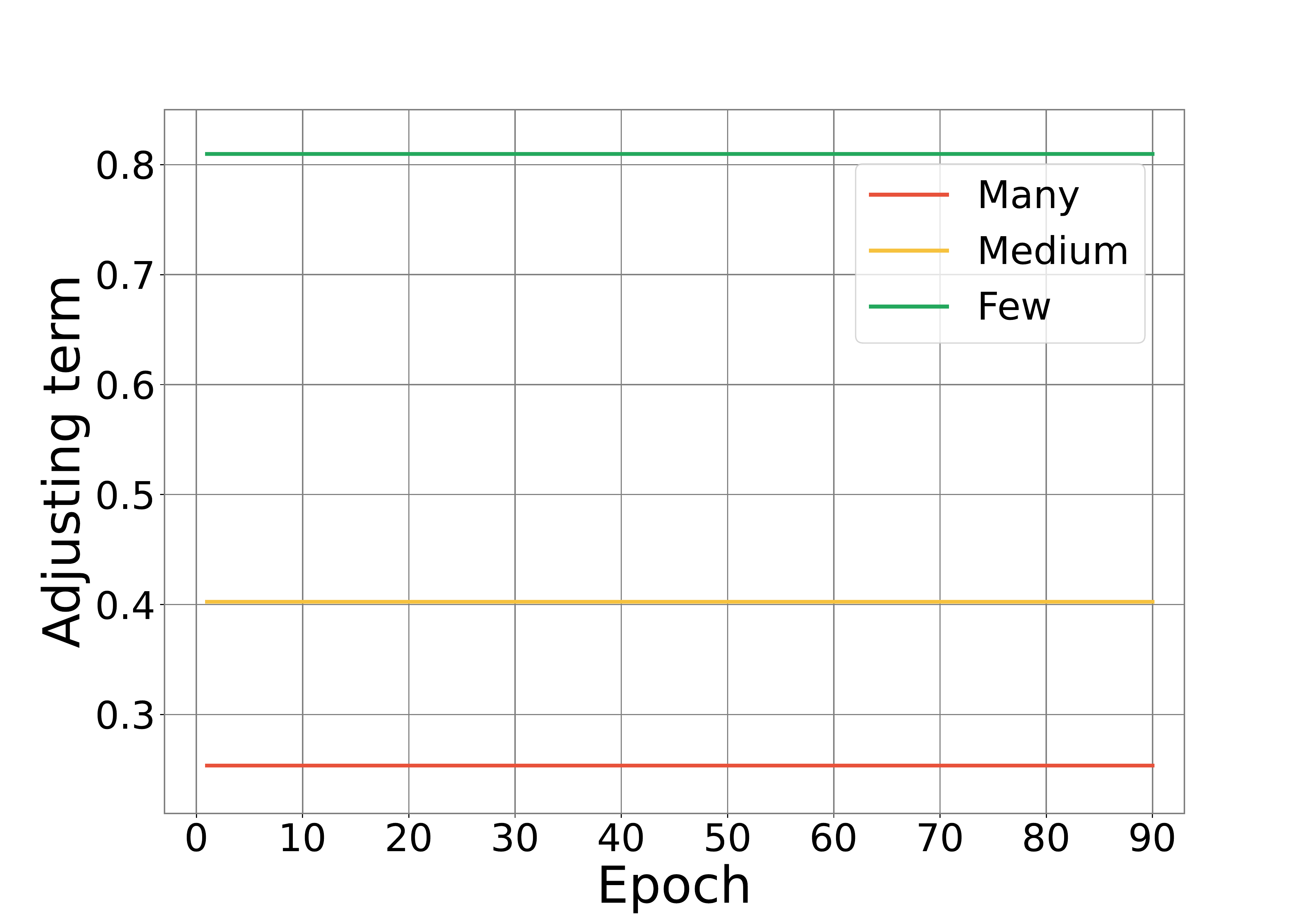}
      \end{minipage}
      \label{fig:exp_adjust_qf}
   }
   \subfigure[$\mathcal{A}^{ALA} = \mathcal{DF} * \mathcal{QF}$]{
      \begin{minipage}[b]{0.3\textwidth}
         \includegraphics[width=1\textwidth]{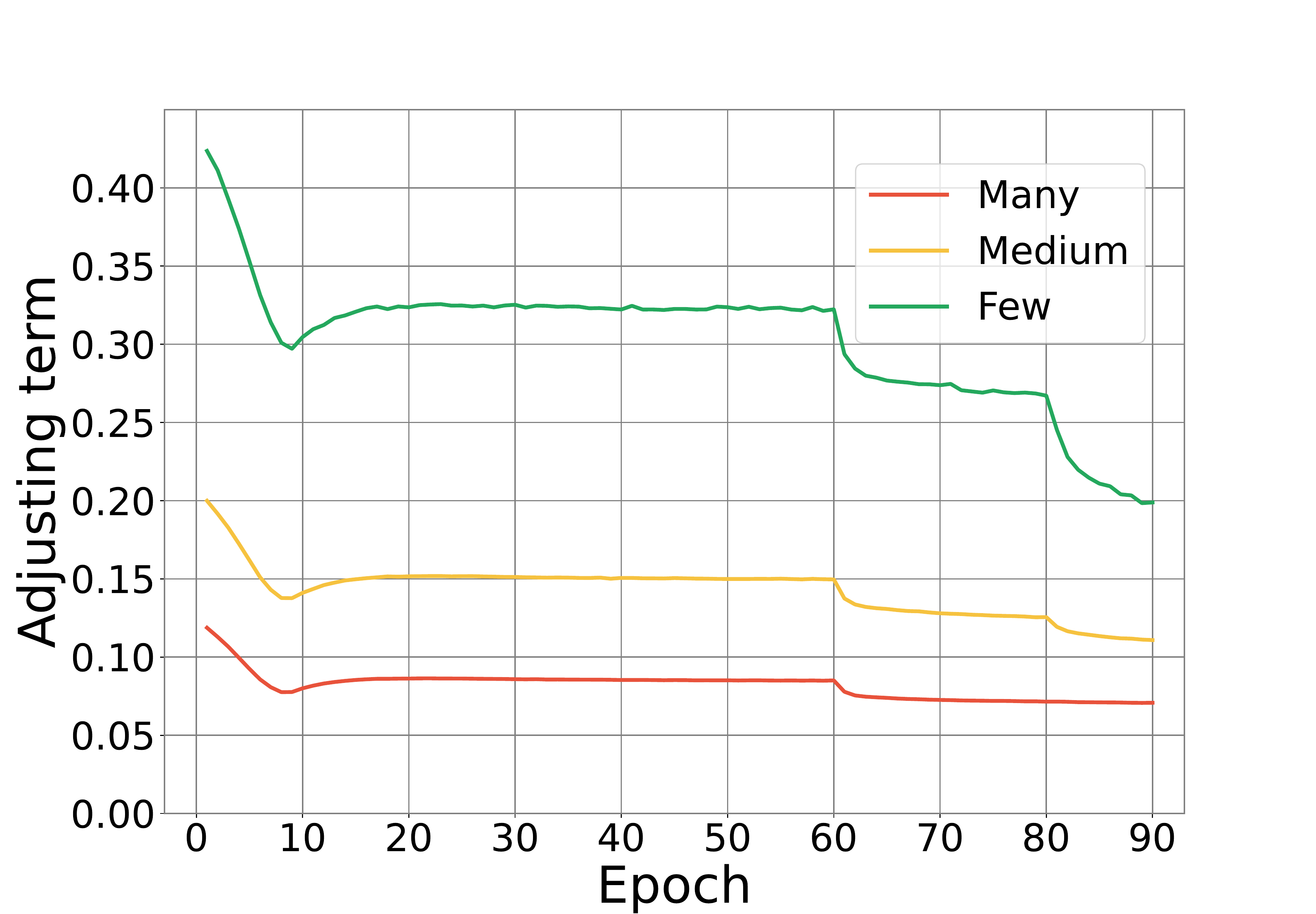}
      \end{minipage}
      \label{fig:exp_adjust_ala}
   }
   \caption{Adjusting curves of $\mathcal{DF}$, $\mathcal{QF}$ and $\mathcal{A}^{ALA}$ in the training process. 
   In order to compare the difference on Many / Medium / Few, the value denotes the average adjusting terms in the same subset.} 
   \label{fig:exp_adjust}
   \vspace{-1em}
\end{figure*}

\subsubsection{Quantitative analysis.}
\label{sec:ablation_quantitative}
In this section, a series of experiments are conducted to examine the effect of each component in ALA Loss. According to the results shown in Table~\ref{tab:exp_ablation}, we have the following observations:

1) $\mathcal{DF}$ aims to optimize hard instances in all subsets, which can boost the performance of both head and tail classes, improving the long-tailed classification to a certain extent. Compared with CE, $\mathcal{DF}$ achieves considerable better results on both medium- and few-shot subsets, with only a slight decline on many-shot subset. Moreover, it even outperforms data quantity based adjusting method LDAM and $\mathcal{QF}$, indicating the advantage of tackling the long-tailed problem from the fine-grained perspective of instance difficulty.

2) $\mathcal{QF}$ performs better on tail classes than LDAM. The comparison between LDAM and $\mathcal{QF}$ reveals that our proposed quantity factor term is able to achieve comparable overall performances as LDAM. What's more, on the few-shot subset, $\mathcal{QF}$ brings significant gains, which is consistent with the analysis in Figure~\ref{fig:method_qf}. 

3) $\mathcal{DF}$ performs better on the many-shot subset, while LDAM and $\mathcal{QF}$ get higher accuracy on the few-shot subset. It is consistent with our design principle. That is: $\mathcal{DF}$ focus more on hard samples, since it tackles the long-tailed problem from the perspective of instance difficulty. $\mathcal{QF}$ and LDAM pay more attention to tail classes, since they are only related with the data quantity.

4) Both $\mathcal{QF}$ and LDAM can be further boosted when combined with our $\mathcal{DF}$. According to the results shown in Table~\ref{tab:exp_ablation}, the combination setting $\mathcal{DF} \cdot \mathcal{QF}$ achieves the best result, and other settings like $\mathcal{DF} \cdot \mathcal{A}^{LDAM}$ also performs better than using either only.

\subsubsection{Qualitative analysis}
In this section, we conduct two qualitative analysis to characterize our ALA Loss intuitively and comprehensively.
Concretely, we further visualize and analyze the advantage of ALA Loss from the perspective of probability and adaptability.

\textit{Probabily analysis.}
To intuitively reflect the advantages of our ALA Loss over data quantity based logit adjusting methods, we conduct an experiment to compare ALA Loss with CE and LDAM.
According to the results shown in Figure~\ref{fig:exp_prob}, we roughly consider those samples with predicted probabilities $< 0.2$ as hard samples, and those $>0.8$ as easy samples. 
It is worth noticing that: 
1) For the medium- and few-shot subsets shown in Figure~\ref{fig:exp_prob_med} and Figure~\ref{fig:exp_prob_few}, both LDAM and ALA Loss can significantly improve the predicted probabilities compared with CE;
2) For the many-shot subset shown in Figure~\ref{fig:exp_prob_many}, LDAM has much more hard samples than CE, which brings significant drop to the accuracy. However, ALA Loss alleviates the excessive hard samples caused by LDAM, with only a slight accuracy decline on many-shot subset compared with CE.
Again, from the perspective of probability, it verifies that our method does not need to excessively sacrifice the performance of head classes in exchange for the promotion of tail classes.

\textit{Adaptability analysis.}
We also visualize the trend of adjusting term in the training process 
in Figure~\ref{fig:exp_adjust}. We discuss it from three aspects:

Firstly, as shown in Figure~\ref{fig:exp_adjust_df}, we obtain the following observations about $\mathcal{DF}$:

1) From head to tail classes, the adjusting term increases gradually, which indicates that the proportion of hard samples in tail is larger than head classes. It can be verified by the ablation experiments in Table~\ref{tab:exp_ablation}: $\mathcal{DF}$ bring significant performance gains for tail classes compared with CE.

2) $\mathcal{DF}$ adaptively changes as the performance of network fluctuates in the training process, and gets smaller with the optimization of the network. 

Secondly, as shown in Figure~\ref{fig:exp_adjust_qf}, $\mathcal{QF}$ keeps unchanged in the whole training process, encouraging larger regularization for tail classes.

Lastly, as shown in Figure~\ref{fig:exp_adjust_ala}, $\mathcal{A}^{ALA}$ adjusts more on the few-shot subset. But the relative difference between the adjusting term of head and tail classes shrinks compared with $\mathcal{QF}$ in Figure~\ref{fig:exp_adjust_qf}, which can alleviate the under-optimization for head yet hard and over-optimization for tail yet easy instances.

%% file: sections/5_conclusion.tex
\section{Conclusion}

In this work, we analyze the issues behind the existing methods for long-tailed classification and propose to revisit this problem from the perspective of not only data quantity but also instance difficulty. 
Our analysis shows that: it is unreasonable for previous logit adjusting methods to simply regularize more on tail classes, which leads to the over-optimization on tail yet easy instances and under-optimization on head yet hard instances.
Therefore, we propose an Adaptive Logit Adjustment (ALA) loss that contains a difficulty factor to focus the model more on hard instances and a quantity factor to make the model pay more attention to tail classes. 
Extensive and comprehensive experimental results show that our method outperforms the existing SOTA methods on three widely used large-scale long-tailed benchmarks including ImageNet-LT, iNaturalist 2018 and Places-LT. 
We will also apply the proposed ALA Loss to the long-tailed object detection and instance segmentation datasets in the future work.

%% file: main.bbl
\begin{thebibliography}{34}
\providecommand{\natexlab}[1]{#1}

\bibitem[{Buda, Maki, and Mazurowski(2018)}]{buda2018systematic}
Buda, M.; Maki, A.; and Mazurowski, M.~A. 2018.
\newblock A systematic study of the class imbalance problem in convolutional
  neural networks.
\newblock \emph{Neural Networks}, 106: 249--259.

\bibitem[{Cao et~al.(2019)Cao, Wei, Gaidon, Arechiga, and Ma}]{cao2019learning}
Cao, K.; Wei, C.; Gaidon, A.; Arechiga, N.; and Ma, T. 2019.
\newblock Learning imbalanced datasets with label-distribution-aware margin
  loss.
\newblock \emph{arXiv preprint arXiv:1906.07413}.

\bibitem[{Chawla et~al.(2002)Chawla, Bowyer, Hall, and
  Kegelmeyer}]{chawla2002smote}
Chawla, N.~V.; Bowyer, K.~W.; Hall, L.~O.; and Kegelmeyer, W.~P. 2002.
\newblock SMOTE: synthetic minority over-sampling technique.
\newblock \emph{Journal of artificial intelligence research}, 16: 321--357.

\bibitem[{Cui et~al.(2019)Cui, Jia, Lin, Song, and Belongie}]{cui2019class}
Cui, Y.; Jia, M.; Lin, T.-Y.; Song, Y.; and Belongie, S. 2019.
\newblock Class-balanced loss based on effective number of samples.
\newblock In \emph{Proceedings of the IEEE/CVF Conference on Computer Vision
  and Pattern Recognition}, 9268--9277.

\bibitem[{Deng et~al.(2009)Deng, Dong, Socher, Li, Li, and
  Fei-Fei}]{deng2009imagenet}
Deng, J.; Dong, W.; Socher, R.; Li, L.-J.; Li, K.; and Fei-Fei, L. 2009.
\newblock Imagenet: A large-scale hierarchical image database.
\newblock In \emph{2009 IEEE conference on computer vision and pattern
  recognition}, 248--255. Ieee.

\bibitem[{Deng et~al.(2019)Deng, Guo, Xue, and Zafeiriou}]{deng2019arcface}
Deng, J.; Guo, J.; Xue, N.; and Zafeiriou, S. 2019.
\newblock Arcface: Additive angular margin loss for deep face recognition.
\newblock In \emph{Proceedings of the IEEE/CVF Conference on Computer Vision
  and Pattern Recognition}, 4690--4699.

\bibitem[{Drumnond(2003)}]{drumnond2003class}
Drumnond, C. 2003.
\newblock Class Imbalance and Cost Sensitivity: Why Undersampling beats
  Oversampling.
\newblock In \emph{ICML-KDD 2003 Workshop: Learning from Imbalanced Datasets}.

\bibitem[{Gupta, Dollar, and Girshick(2019)}]{gupta2019lvis}
Gupta, A.; Dollar, P.; and Girshick, R. 2019.
\newblock {LVIS}: A Dataset for Large Vocabulary Instance Segmentation.
\newblock In \emph{Proceedings of the {IEEE} Conference on Computer Vision and
  Pattern Recognition}.

\bibitem[{Han, Wang, and Mao(2005)}]{han2005borderline}
Han, H.; Wang, W.-Y.; and Mao, B.-H. 2005.
\newblock Borderline-SMOTE: a new over-sampling method in imbalanced data sets
  learning.
\newblock In \emph{International conference on intelligent computing},
  878--887. Springer.

\bibitem[{He et~al.(2016)He, Zhang, Ren, and Sun}]{he2016deep}
He, K.; Zhang, X.; Ren, S.; and Sun, J. 2016.
\newblock Deep residual learning for image recognition.
\newblock In \emph{Proceedings of the IEEE conference on computer vision and
  pattern recognition}, 770--778.

\bibitem[{Hong et~al.(2021)Hong, Han, Choi, Seo, Kim, and
  Chang}]{hong2021disentangling}
Hong, Y.; Han, S.; Choi, K.; Seo, S.; Kim, B.; and Chang, B. 2021.
\newblock Disentangling Label Distribution for Long-tailed Visual Recognition.
\newblock In \emph{Proceedings of the IEEE/CVF Conference on Computer Vision
  and Pattern Recognition}, 6626--6636.

\bibitem[{Kang et~al.(2019)Kang, Xie, Rohrbach, Yan, Gordo, Feng, and
  Kalantidis}]{kang2019decoupling}
Kang, B.; Xie, S.; Rohrbach, M.; Yan, Z.; Gordo, A.; Feng, J.; and Kalantidis,
  Y. 2019.
\newblock Decoupling representation and classifier for long-tailed recognition.
\newblock \emph{arXiv preprint arXiv:1910.09217}.

\bibitem[{Khan et~al.(2017)Khan, Hayat, Bennamoun, Sohel, and
  Togneri}]{khan2017cost}
Khan, S.~H.; Hayat, M.; Bennamoun, M.; Sohel, F.~A.; and Togneri, R. 2017.
\newblock Cost-sensitive learning of deep feature representations from
  imbalanced data.
\newblock \emph{IEEE transactions on neural networks and learning systems},
  29(8): 3573--3587.

\bibitem[{Lin et~al.(2017)Lin, Goyal, Girshick, He, and
  Doll{\'a}r}]{lin2017focal}
Lin, T.-Y.; Goyal, P.; Girshick, R.; He, K.; and Doll{\'a}r, P. 2017.
\newblock Focal loss for dense object detection.
\newblock In \emph{Proceedings of the IEEE international conference on computer
  vision}, 2980--2988.

\bibitem[{Lin et~al.(2014)Lin, Maire, Belongie, Hays, Perona, Ramanan,
  Doll{\'a}r, and Zitnick}]{lin2014microsoft}
Lin, T.-Y.; Maire, M.; Belongie, S.; Hays, J.; Perona, P.; Ramanan, D.;
  Doll{\'a}r, P.; and Zitnick, C.~L. 2014.
\newblock Microsoft coco: Common objects in context.
\newblock In \emph{European conference on computer vision}, 740--755. Springer.

\bibitem[{Liu et~al.(2017)Liu, Wen, Yu, Li, Raj, and Song}]{liu2017sphereface}
Liu, W.; Wen, Y.; Yu, Z.; Li, M.; Raj, B.; and Song, L. 2017.
\newblock Sphereface: Deep hypersphere embedding for face recognition.
\newblock In \emph{Proceedings of the IEEE conference on computer vision and
  pattern recognition}, 212--220.

\bibitem[{Liu et~al.(2016)Liu, Wen, Yu, and Yang}]{liu2016large}
Liu, W.; Wen, Y.; Yu, Z.; and Yang, M. 2016.
\newblock Large-margin softmax loss for convolutional neural networks.
\newblock In \emph{ICML}, volume~2, 7.

\bibitem[{Liu et~al.(2019)Liu, Miao, Zhan, Wang, Gong, and Yu}]{liu2019large}
Liu, Z.; Miao, Z.; Zhan, X.; Wang, J.; Gong, B.; and Yu, S.~X. 2019.
\newblock Large-scale long-tailed recognition in an open world.
\newblock In \emph{Proceedings of the IEEE/CVF Conference on Computer Vision
  and Pattern Recognition}, 2537--2546.

\bibitem[{Menon et~al.(2013)Menon, Narasimhan, Agarwal, and
  Chawla}]{menon2013statistical}
Menon, A.; Narasimhan, H.; Agarwal, S.; and Chawla, S. 2013.
\newblock On the statistical consistency of algorithms for binary
  classification under class imbalance.
\newblock In \emph{International Conference on Machine Learning}, 603--611.
  PMLR.

\bibitem[{Menon et~al.(2020)Menon, Jayasumana, Rawat, Jain, Veit, and
  Kumar}]{menon2020long}
Menon, A.~K.; Jayasumana, S.; Rawat, A.~S.; Jain, H.; Veit, A.; and Kumar, S.
  2020.
\newblock Long-tail learning via logit adjustment.
\newblock \emph{arXiv preprint arXiv:2007.07314}.

\bibitem[{Ren et~al.(2018)Ren, Zeng, Yang, and Urtasun}]{ren2018learning}
Ren, M.; Zeng, W.; Yang, B.; and Urtasun, R. 2018.
\newblock Learning to reweight examples for robust deep learning.
\newblock In \emph{International Conference on Machine Learning}, 4334--4343.
  PMLR.

\bibitem[{Russakovsky et~al.(2015)Russakovsky, Deng, Su, Krause, Satheesh, Ma,
  Huang, Karpathy, Khosla, Bernstein et~al.}]{russakovsky2015imagenet}
Russakovsky, O.; Deng, J.; Su, H.; Krause, J.; Satheesh, S.; Ma, S.; Huang, Z.;
  Karpathy, A.; Khosla, A.; Bernstein, M.; et~al. 2015.
\newblock Imagenet large scale visual recognition challenge.
\newblock \emph{International journal of computer vision}, 115(3): 211--252.

\bibitem[{Shu et~al.(2019)Shu, Xie, Yi, Zhao, Zhou, Xu, and Meng}]{shu2019meta}
Shu, J.; Xie, Q.; Yi, L.; Zhao, Q.; Zhou, S.; Xu, Z.; and Meng, D. 2019.
\newblock Meta-weight-net: Learning an explicit mapping for sample weighting.
\newblock \emph{arXiv preprint arXiv:1902.07379}.

\bibitem[{Tan et~al.(2020)Tan, Wang, Li, Li, Ouyang, Yin, and
  Yan}]{tan2020equalization}
Tan, J.; Wang, C.; Li, B.; Li, Q.; Ouyang, W.; Yin, C.; and Yan, J. 2020.
\newblock Equalization loss for long-tailed object recognition.
\newblock In \emph{Proceedings of the IEEE/CVF Conference on Computer Vision
  and Pattern Recognition}, 11662--11671.

\bibitem[{Van~Horn et~al.(2018)Van~Horn, Mac~Aodha, Song, Cui, Sun, Shepard,
  Adam, Perona, and Belongie}]{van2018inaturalist}
Van~Horn, G.; Mac~Aodha, O.; Song, Y.; Cui, Y.; Sun, C.; Shepard, A.; Adam, H.;
  Perona, P.; and Belongie, S. 2018.
\newblock The inaturalist species classification and detection dataset.
\newblock In \emph{Proceedings of the IEEE conference on computer vision and
  pattern recognition}, 8769--8778.

\bibitem[{Wallace et~al.(2011)Wallace, Small, Brodley, and
  Trikalinos}]{wallace2011class}
Wallace, B.~C.; Small, K.; Brodley, C.~E.; and Trikalinos, T.~A. 2011.
\newblock Class imbalance, redux.
\newblock In \emph{2011 IEEE 11th international conference on data mining},
  754--763. IEEE.

\bibitem[{Wang et~al.(2018{\natexlab{a}})Wang, Cheng, Liu, and
  Liu}]{wang2018additive}
Wang, F.; Cheng, J.; Liu, W.; and Liu, H. 2018{\natexlab{a}}.
\newblock Additive margin softmax for face verification.
\newblock \emph{IEEE Signal Processing Letters}, 25(7): 926--930.

\bibitem[{Wang et~al.(2018{\natexlab{b}})Wang, Wang, Zhou, Ji, Gong, Zhou, Li,
  and Liu}]{wang2018cosface}
Wang, H.; Wang, Y.; Zhou, Z.; Ji, X.; Gong, D.; Zhou, J.; Li, Z.; and Liu, W.
  2018{\natexlab{b}}.
\newblock Cosface: Large margin cosine loss for deep face recognition.
\newblock In \emph{Proceedings of the IEEE conference on computer vision and
  pattern recognition}, 5265--5274.

\bibitem[{Wang et~al.(2020)Wang, Lian, Miao, Liu, and Yu}]{wang2020long}
Wang, X.; Lian, L.; Miao, Z.; Liu, Z.; and Yu, S.~X. 2020.
\newblock Long-tailed Recognition by Routing Diverse Distribution-Aware
  Experts.
\newblock \emph{arXiv preprint arXiv:2010.01809}.

\bibitem[{Xiang, Ding, and Han(2020)}]{xiang2020learning}
Xiang, L.; Ding, G.; and Han, J. 2020.
\newblock Learning from multiple experts: Self-paced knowledge distillation for
  long-tailed classification.
\newblock In \emph{European Conference on Computer Vision}, 247--263. Springer.

\bibitem[{Yin et~al.(2019)Yin, Yu, Sohn, Liu, and Chandraker}]{yin2019feature}
Yin, X.; Yu, X.; Sohn, K.; Liu, X.; and Chandraker, M. 2019.
\newblock Feature transfer learning for face recognition with under-represented
  data.
\newblock In \emph{Proceedings of the IEEE/CVF Conference on Computer Vision
  and Pattern Recognition}, 5704--5713.

\bibitem[{Zhang et~al.(2021)Zhang, Li, Yan, He, and
  Sun}]{zhang2021distribution}
Zhang, S.; Li, Z.; Yan, S.; He, X.; and Sun, J. 2021.
\newblock Distribution Alignment: A Unified Framework for Long-tail Visual
  Recognition.
\newblock In \emph{Proceedings of the IEEE/CVF Conference on Computer Vision
  and Pattern Recognition}, 2361--2370.

\bibitem[{Zhou et~al.(2020)Zhou, Cui, Wei, and Chen}]{zhou2020bbn}
Zhou, B.; Cui, Q.; Wei, X.-S.; and Chen, Z.-M. 2020.
\newblock Bbn: Bilateral-branch network with cumulative learning for
  long-tailed visual recognition.
\newblock In \emph{Proceedings of the IEEE/CVF Conference on Computer Vision
  and Pattern Recognition}, 9719--9728.

\bibitem[{Zhou et~al.(2017)Zhou, Lapedriza, Khosla, Oliva, and
  Torralba}]{zhou2017places}
Zhou, B.; Lapedriza, A.; Khosla, A.; Oliva, A.; and Torralba, A. 2017.
\newblock Places: A 10 million image database for scene recognition.
\newblock \emph{IEEE transactions on pattern analysis and machine
  intelligence}, 40(6): 1452--1464.

\end{thebibliography}
